\documentclass{article}
\usepackage{ijcai19}

\usepackage{times}
\usepackage{soul}
\usepackage{url}
\usepackage{hyperref}
\usepackage[utf8]{inputenc}
\usepackage[small]{caption}
\usepackage{graphicx}
\usepackage{amsmath}
\usepackage{booktabs}
\usepackage{algorithm}
\usepackage{algorithmic}
\renewcommand{\algorithmiccomment}[1]{\bgroup\hfill//~#1\egroup}
\usepackage[subrefformat=parens,format=hang]{subcaption} 

\title{Macro Action Reinforcement Learning with Sequence Disentanglement using Variational Autoencoder}

\author{
Heecheol Kim$^1$\footnote{Both authors equally contributed to this paper.}\and
Masanori Yamada$^2$\footnotemark[1]\and
Kosuke Miyoshi$^{3}$\And
Hiroshi Yamakawa$^4$
\affiliations
$^{1,3,4}$Dwango Artificial Intelligence Laboratory\\
$^2$NTT Secure Platform Laboratories\
\emails
$^1$h-kim@isi.imi.i.u-tokyo.ac.jp\\
$^2$masanori.yamada.cm@hco.ntt.co.jp\\
$^3$miyoshi@narr.jp\\
$^4$hiroshi\_yamakawa@dwango.co.jp
}

\begin{document}

\maketitle

\begin{abstract}
One problem in the application of reinforcement learning to real-world problems is the curse of dimensionality on the action space.
Macro actions, a sequence of primitive actions, have been studied to diminish the dimensionality of the action space with regard to the time axis.
However, previous studies relied on humans defining macro actions or assumed macro actions as repetitions of the same primitive actions.
We present Factorized Macro Action Reinforcement Learning (FaMARL) which autonomously learns disentangled factor representation of a sequence of actions to generate macro actions that can be directly applied to general reinforcement learning algorithms.
FaMARL exhibits higher scores than other reinforcement learning algorithms on environments that require an extensive amount of search.
\end{abstract}

\section{Introduction}\label{Intro}

Reinforcement learning has gained significant attention recently in both robotics and machine-learning communities because of its potential of wide application to different domains. Recent studies have achieved above-human level game play in Go \cite{silver2016mastering,silver2017mastering} and video games \cite{mnih2015human,OpenAI_dota}. Application of reinforcement learning to real-world robots has also been widely studied \cite{levine2016end,haarnoja2018learning}.

Reinforcement learning involves learning the relationship between a state and action on the basis of rewards. Reinforcement learning fails when the dimensionality of a state or action increases. This is why reinforcement learning is often considered data inefficient, i.e., requiring a large number of trials. 
The curse of dimensionality on the state space is partially solved using a convolutional neural network (CNN) \cite{krizhevsky2012imagenet,mnih2015human}; training policy from raw image input has become possible by applying a CNN against the input states.
However, reducing the dimensionality on the action side is still challenging. The search space can be exponentially wide with a longer sequence and higher action dimension.

Application of macro actions to reinforcement learning has been studied to reduce the dimensionality of actions. By compressing the sequence of primitive actions, macro actions diminish the search space.
Previous studies defined macro actions as repetitions of the same primitive actions \cite{sharma2017learning} or requiring humans to manually define them \cite{hausknecht2015deep}. However, more sophisticated macro actions should contain different primitive actions in one sequence without humans having to manually defining these actions. 

We propose Factorized Macro Action Reinforcement Learning (FaMARL), a novel algorithm for abstracting the sequence of primitive actions to macro actions by learning disentangled representation \cite{bengio2013deep} of a given sequence of actions, reducing dimensionality of the action search space. Our algorithm uses Factorized Action Variational Autoencoder (FAVAE) \cite{yamada2019favae}, a variation of VAE \cite{kingma2013auto}, to learn macro actions from given expert demonstrations. Using the acquired disentangled latent variables as macro actions, FaMARL matches the state with the latent variables of FAVAE instead of primitive actions directly. The matched latent variables are then decoded into a sequence of primitive actions and applied repeatedly to the environment. FaMARL is not limited to just repeating the same primitive actions multiple times, because this compresses any kind of representation with FAVAE. We experimentally show that FaMARL can learn environments with high dimensionality of the search space.

\section{Related work}


Applying a sequence of actions to reinforcement learning has been studied \cite{sharma2017learning,vezhnevets2016strategic,lakshminarayanan2017dynamic,durugkar2016deep}. Fine Grained Action Repetition (FiGAR) successfully adopts macro actions into deep reinforcement learning \cite{sharma2017learning}, showing that Asynchronous Advantage Actor-Critic (A3C)\cite{mnih2016asynchronous}, an asynchronous variant of deep reinforcement learning algorithm, with a learning time scale of repeating the action as well as the action itself scores higher than that with primitive actions in Atari 2600 Games.

There are mainly two differences between FaMARL and FiGAR. 
First, FiGAR can only generate macro actions that are the repeat of the same primitive actions. On the other hand, macro actions generated with FaMARL can be a combination of different primitive actions because FaMARL finds a disentangled representation of a sequence of continuous actions and uses the decoded sequence as macro actions.
Second, FaMARL learns how to generate macro actions and optimizes the policy for the target task independently, while FiGAR learns both simultaneously. Despite FaMARL cannot learn macro actions end-to-end, this algorithm can easily recycle acquired macro actions to new target tasks, because macro actions are acquired independent to target tasks.

Hausknecht proposed using a parameterized continuous action space in the reinforcement learning framework \cite{hausknecht2015deep}. This approach, however, is limited in the fact that the action has to be selected at every time step, and humans need to parameterize the action. FaMARL can be viewed as an expansion of this model to time series.







\section{Sequence-Disentanglement Representation Learning by Factorized Action Variational AutoEncoder}

VAE \cite{kingma2013auto} is a generative model that learns probabilistic latent variables $z$ via the probability distribution learning of a dataset. VAE encodes data $x$ to latent variable $z$ and reconstructs $x$ from $z$.

The $\beta$-VAE \cite{higgins2017beta} and CCI-VAE \cite{burgess2018understanding}, which is an improved $\beta$-VAE, are models for learning the disentangled representations. These models disentangle $z$  by adding the constraint to reduce the total correlation to VAE. FAVAE \cite{yamada2019favae} is an extended $\beta$-VAE model to learn disentangled representations from sequential data. FAVAE has a ladder network structure and information-bottleneck-type loss function. This loss function of FAVAE is defined as 
\begin{align}
&-E_{q_{\phi}\left(z|\left(x_{1:T}\right)\right)}\left[\log p_{\theta}\left(x_{1:T}|z\right)\right]\nonumber\\
&+\beta\sum_{\tilde{l}}\left|D_{{\rm KL}}\left(q_{\phi}\left(z|\left(x_{1:T}\right)\right)||p\left(z\right)\right)_{\tilde{l}}-C_{\tilde{l}}\right|,\label{eq:beta_vae}
\end{align}
where $p\left(z\right)=\mathcal{N}\left(0,1\right)$, $\tilde{l}$ is the index of the ladder, $\beta$ is a constant greater than zero that encourages disentangled representation learning by weighting
Kullback-Leibler divergence term, and $C$ is called information capacity for supporting the reconstruction. In the learning phase, $C$ increases linearly along with epochs from $0$ to $C_{last}$. The $C_{last}$ is determined by first training FAVAE with a small amount of $\beta$ (we used $\beta=0.1$) and $C_{last}=0$. The last value of $D_{kl}(q(z|x) || p(z))$ is used as $C_{last}$. Each ladder requires a C . For example, a 3-ladder network requires 3 Cs.





\section{Proposed algorithm}

\begin{figure*}
  \centering \includegraphics[width=0.8\linewidth]{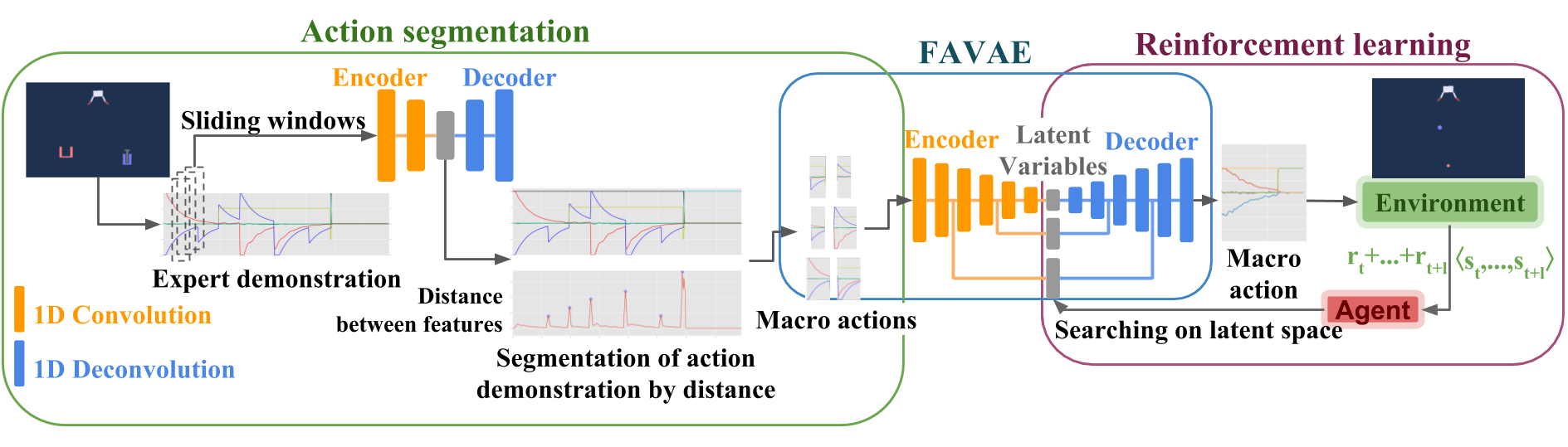}
  \caption{Overview of FaMARL}
  \label{fig:overview}
\end{figure*}


Our objective is to find factorized macro actions from given time series of expert demonstrations and search for the optimal policy of a target task based on these macro actions instead of primitive actions. The target task can differ from the task that the expert demonstrations are generated.
We use FAVAE \cite{yamada2019favae} to find factorized macro actions. The details of FaMARL are given in Sections~\ref{subsection:segmentation} and \ref{subsection:VAE}.



One might be curious why we do not apply expert demonstrations or their segmentations, directly to the reinforcement learning agent to learn a new task. There are two reasons for learning disentangled factors of (segmented) expert demonstrations. 
First, if the agent explores these expert demonstrations only, it can only mimic expert demonstrations to solve the task, which results in serious deficiencies in generalizing macro actions. Consider a set of demonstrations containing actions of $\langle$ turn right $70^\circ$, turn right $60^\circ$, …, turn right $10 ^\circ$ $\rangle$. If the environment requires the agent to turn right $80^\circ$, the agent cannot complete the task. On the other hand, latent variables trained with the expert demonstrations acquire generated macro actions to "\textit{turn right $x^\circ$}. Thus, the agent can easily adapt to the target task.
Second, without latent variables, the action space is composed by listing only all expert demonstrations, forming a discrete action space. This causes the \textit{curse of dimensionality}, detering fast convergence on the task.

\begin{algorithm}[tb]
\caption{Unsupervised segmentation of macro actions}
\label{alg:segmentation}
\textbf{Input}: Expert demonstration $D \gets \langle A_1,A_2,...,A_n \rangle $ on Base, where $A_i \gets \langle a_{i1}, a_{i2}, ..., a_{im} \rangle$ ($m$ $\gets$ length of $i$th episode)\\
\textbf{Parameter}: Encoder $q_{seg}$\\
\begin{algorithmic}[1] 
\STATE $D_{slice} \gets \langle \langle d_{11}, d_{12}, ... \rangle ,  \langle d_{21}, d_{22}, ... \rangle , ..., \langle d_{n1}, d_{n2}, ... \rangle \rangle$ \COMMENT{Slice all $A_i$ with WindowSize $\gets$ 4}
\STATE Train $q_{seg}(d_{ij})$ with $d_{ij} \ni D_{slice}$
\STATE $distance_{ij} \gets |q_{seg}(d_{ij}) - q_{seg}(d_{ij-1})|$
\STATE Segment $A_i \ni D$ with $distance_i$
\STATE $D_{seg} \gets \langle \langle x_{11}, x_{12}, ... \rangle , \langle x_{21}, x_{22}, ... \rangle , ..., \langle x_{n1}, x_{n2}, ... \rangle \rangle$
\end{algorithmic}
\end{algorithm}

\begin{algorithm}[tb]
\caption{Factorized macro action with proximal policy optimization (PPO)}
\label{alg:FaMARL}
\textbf{Input}: Decoder of FAVAE $\theta$\\
\textbf{Parameter}: PPO Agent $\psi$\\
\begin{algorithmic}[1] 
\WHILE{converge}
\STATE $z_t \sim \pi_{\psi} (z_t|s_t)$
\STATE $\langle a_t,a_{t+1},...,a_{t+l} \rangle \gets p_{\theta}(b_t|z_t)$
\STATE $r_{tot} \gets 0$
\FOR{$k \gets t$ to ${t+l}$}
\STATE $s_{k+1},r_k \gets p(a_k, s_k)$
\STATE $r_{tot} \gets r_{tot} + r_k$
\ENDFOR\\
\STATE Minimize equation~\ref{eq:PPO_macro} using $r_{tot}$
\ENDWHILE
\end{algorithmic}
\end{algorithm}

\subsection{Unsupervised segmentation of macro actions}\label{subsection:segmentation}

An episode of an expert demonstration is composed of a series of macro actions, e.g., when humans show a demonstration of moving an object by hand, that demonstration is composed of 1)extending a hand to the object, 2)grasping the object, 3)moving the hand to the target position, and 4)releasing the object.

Therefore, expert demonstrations first need to be segmented into each macro action. One significant challenge is that there are usually no ground-truth labels for macro actions. One possible solution is to ask experts to label their actions. However, this is another burden and incurs additional cost.


Lee proposed a simple method using an AE \cite{hinton2006reducing,vincent2008extracting} to segment signal data \cite{lee2018time}. This method, simply speaking, trains an AE with sliding windows of signal data, acquiring the temporal characteristics of the sliding windows. Then, the distance between the encoded features of two adjacent sliding windows is calculated. All the peaks of the distance curve are selected as segmentation points. One advantage of this method is that it is not domain-specific. This method can be easily applied to expert demonstration data since it is assumed that there are no specific data characteristics.

On our implementation of this segmentation method, distance is defined as $|q_{seg}(a_i)-q_{seg}(a_{i-1})|$, where $q_{seg}(a_{ij})$ refers to the encoded feature of the $j$th sliding window on $i$th trajectory data. We used a sliding window size of $4$. Any distance point that is highest among $10$ adjacent points with a margin of $0.05$ is selected as a peak.





\subsection{Learning disentangled latent variables with FAVAE}\label{subsection:VAE}

Once the expert demonstrations are segmented, FAVAE learns factors that compose macro actions. However, FAVAE cannot directly intake segmented macro actions. This is because segmented macro actions may have different lengths, while FAVAE cannot compute data with different lengths because it uses a combination of 1D convolution and multilayer perceptron which requires an unified data size across all datasets. To address this issue, macro actions are padded with trailing zeros to match the data length of $L$, the input size of FAVAE. Also, two additional dimensions $action_{on}$ and $action_{off}$ are added to macro actions to identify if action $a_k$ at timestep $k$ is a real action or zero-padded one. The $action_{on}$ is $\langle 1_0,1_1,1_2,...1_l,0_{l+1},0_{l+2},...0_L \rangle$ and $action_{off}$ is $\langle 0_0,0_1,0_2,...0_l,1_{l+1},1_{l+2},...1_L \rangle$, where subscript $l$ is the length of a macro action and subscript $L$ is the input size of FAVAE. The cutting point of a real action against zero-padding is computed by the first timestep where $action_{off}$ is selected from the softmax of $action_{on}$ and $action_{off}$.
We used the mean squared error for reconstruction loss. Also, FAVAE used three ladders and CCI is applied. \cite{burgess2018understanding}.




\subsection{Learning policy with proximal policy optimization (PPO)}\label{subsection:PPO}
Our key idea of diminishing the search space is to search on the latent space of the macro actions instead of primitive actions directly. We used proximal policy optimization (PPO) \cite{schulman2017proximal} as the reinforcement learning algorithm, although any kind of reinforcement learning algorithm can be used\footnote{Our implementation of PPO is based on \href{https://github.com/Anjum48/rl-examples}{https://github.com/Anjum48/rl-examples}}. 

PPO is used following the loss function:
\begin{equation}\label{eq:PPO}
\resizebox{.91\linewidth}{!}{$
    \displaystyle
L^{CLIP}(\psi)=E_t[\min(\rho_t(\psi)\hat{A_t},\text{clip} (\rho_t(\psi),1-\epsilon,1+\epsilon)\hat{A_t})
$}
\end{equation}

Here, $\rho_t(\psi)=\frac{\pi_{\psi}(a_t|s_t)}{\pi_{\text{old}}(a_t|s_t)}$, where $\rho_t$ denotes the probability ratio.

Integrating PPO with macro actions generated with FAVAE is simply to replace the primitive action of every time step with the macro action with a step interval $l$ which is the length of the macro action. Therefore, the model of the environment with respect to a macro action is:

\begin{equation}\label{eq:state_transfer}
    s_{t+l}, \sum_{k=t}^{t+l} r_k \sim p(z_t,s_t)
\end{equation}
where $p(z_t,s_t)$ is the transition model of the environment.

The PPO agent matches a latent variable $z_t$ on input state $s_t$. 

The decoder $\phi$ of FAVAE then decodes $z_t$ into series of actions: $ \langle a_t, a_{t+1}, a_{t+2}, ..., a_{t+L} \rangle$, where subscript $L$ is the output length of the decoder. Then actions are trimmed using the value of the softmax of $action_{on}$ and $action_{off}$, which is also decoded from the decoder. 

The macro action is cropped to $\langle a_t, a_{t+1}, ..., a_{t+l} \rangle$  where subscript $l$ is the first timestep at which $action_{off}$ is selected.
This macro action is applied to the environment without feedback. Rewards between $t$ and $t+l$ are summed and regarded as the reward for the result of output $z_{t+l}$.

Thus, the objective function of PPO can be modified as:

\begin{equation}\label{eq:PPO_macro}
\resizebox{.91\linewidth}{!}{$
    \displaystyle
L^{CLIP}(\psi)=E_{t'}[\min(\rho_{t'}(\psi)\hat{A_{t'}},\text{clip} (\rho_{t'}(\psi),1-\epsilon,1+\epsilon)\hat{A_{t'}})    
$}
\end{equation}

where $t'$ is the time step from the perspective of the macro action. If $t$ and $t'$ indicate the same time step in the environment, the relationship of $t+l = t'+1$ is established.

\section{Experiments}

FaMARL was tested in two environments: ContinuousWorld, a simple 2D environment with continuous action and state spaces, and RobotHand, a 2D environment with simulated robot hand made by Box2D, a 2D physics engine\footnote{Dataset and other supplementary results are available at \href{https://github.com/FaMARLSupplement/FaMARLSupplement}{https://github.com/FaMARLSupplement/FaMARLSupplement}}.

\subsection{ContinuousWorld}\label{sec:cont_world}
\begin{figure}
    \centering
    \begin{minipage}{0.35\hsize}
        \includegraphics[width=\linewidth]{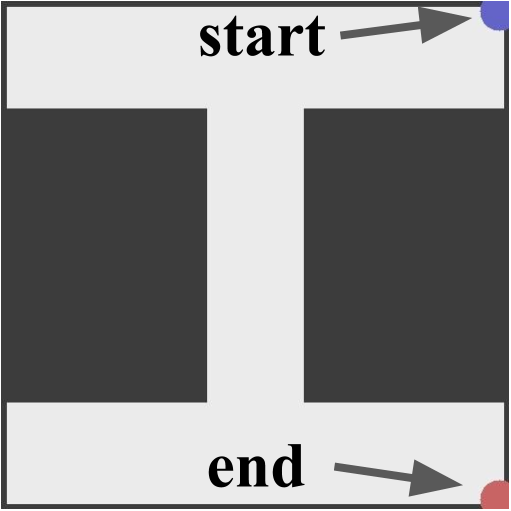}
        \subcaption{Base}
        \label{fig:cwmaze}
    \end{minipage}
    \begin{minipage}{0.35\hsize}
        \includegraphics[width=\linewidth]{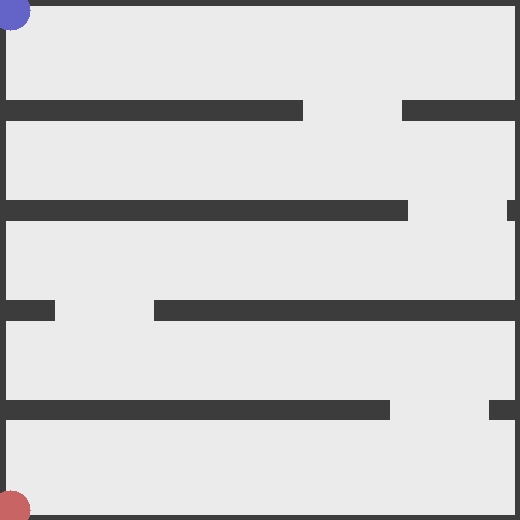}
        \subcaption{Maze}
        \label{fig:cwkey}
    \end{minipage}
    \caption{ContinuousWorld tasks} \label{fig:continuous_world}
\end{figure}

The objective with this environment is to find the optimal trajectory from the starting position (blue dot in Figure ~\ref{fig:continuous_world}) to the goal position (red dot in Figure ~\ref{fig:continuous_world}). The reward of this environment is $-|x-g|$, where $x$ is the position of the agent and $g$ is the position of the goal. The action space is defined by the $\langle$ acceleration to the $x$ axis and acceleration to the $y$ axis $\rangle$. 



There are two tasks in ContinuousWorld: Base and Maze.
In Base, the agent and goal are randomly placed at the corners, top or bottom. Thus, the number of cases for initialization is $2*2=4$. To acquire factors of macro actions regardless of scale, the size of map is selected between $[2.5,5]$ randomly.
In Maze, the agent and goal are always placed at the same position. However, the entrances in the four walls are set randomly for each episode so that the agent has to find an optimal policy on different entrance positions. This makes this environment difficult because walls act like strong local optima of reward; the agent has to make a long detour with lower rewards to finally find the optimal policy.

\begin{figure}
\centering
    \begin{minipage}{0.2\hsize}
        \includegraphics[width=\linewidth]{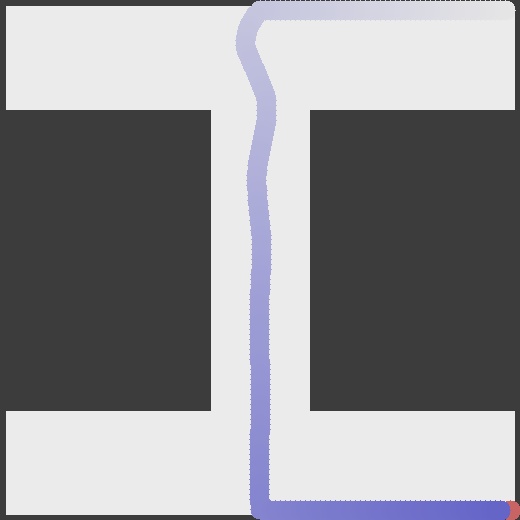}
        \subcaption{}
        \label{fig:down}
    \end{minipage}
    \begin{minipage}{0.2\hsize}
        \includegraphics[width=\linewidth]{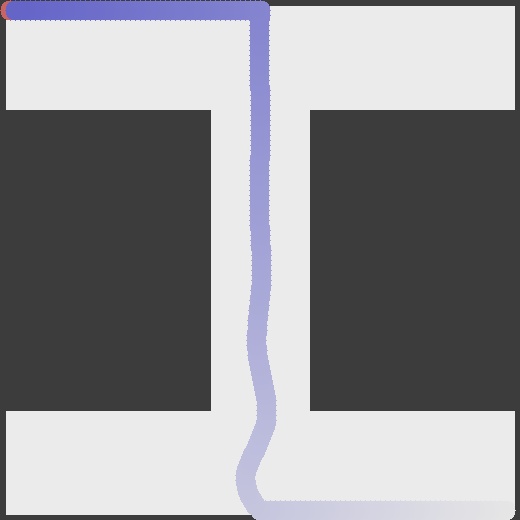}
        \subcaption{}
        \label{fig:up}
    \end{minipage}
    \begin{minipage}{0.2\hsize}
        \includegraphics[width=\linewidth]{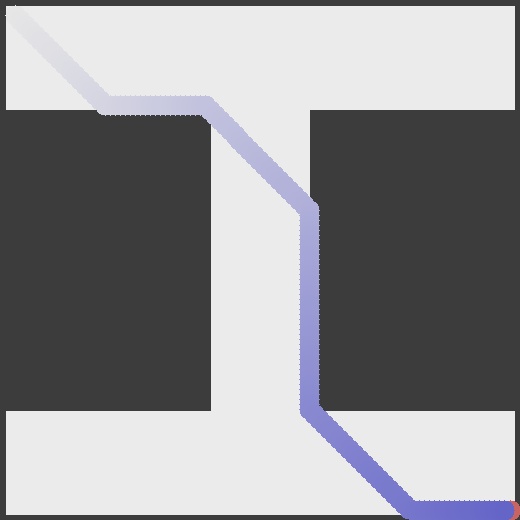}
        \subcaption{}
        \label{fig:pushed_down}
    \end{minipage}
    \begin{minipage}{0.2\hsize}
        \includegraphics[width=\linewidth]{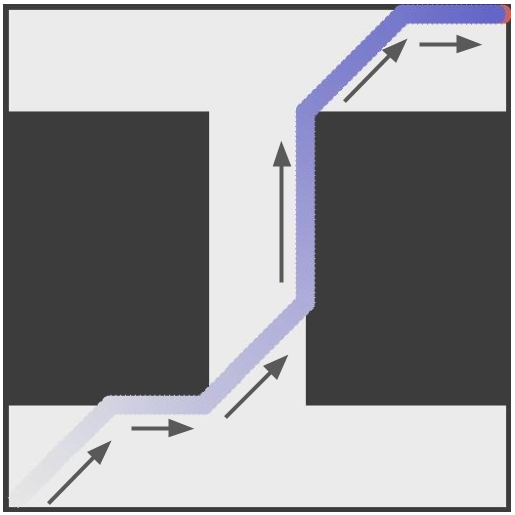}
        \subcaption{}
        \label{fig:pushed_up}
    \end{minipage}
    \caption{Examples of script trajectories. DownOnly uses only trajectories in \ref{fig:down}, Down\&Up uses those in \ref{fig:down} and \ref{fig:up}, PushedDownOnly uses those in \ref{fig:pushed_down}, and PushedDown\&Up uses those in \ref{fig:pushed_down} and \ref{fig:pushed_up}} \label{fig:4 scripts}
\end{figure}


Our purpose was to find disentangled macro actions from expert demonstrations in Base and applying the macro actions to complete the target tasks.
100 episodes of the expert demonstrations were generated in Base using programmed scripts. We compared four different scripts: DownOnly, Down\&Up, PushedDownOnly, and PushedDown\&Up. All scripts are illustrated in Figure~\ref{fig:4 scripts}. For DownOnly, the goal is only initialized at the bottom of the aisle; therefore, the macro actions do not include upward movements. On the other hand, Down\&Up does not limit the position of the goal; thus, upward and downward movements are included in the macro actions. For PushedDownOnly and PushedDown\&Up, the agent always accelerates upward or downward, according to the goal position.


\begin{figure}
    \centering
    \begin{minipage}{0.8\hsize}
        \includegraphics[width=\linewidth]{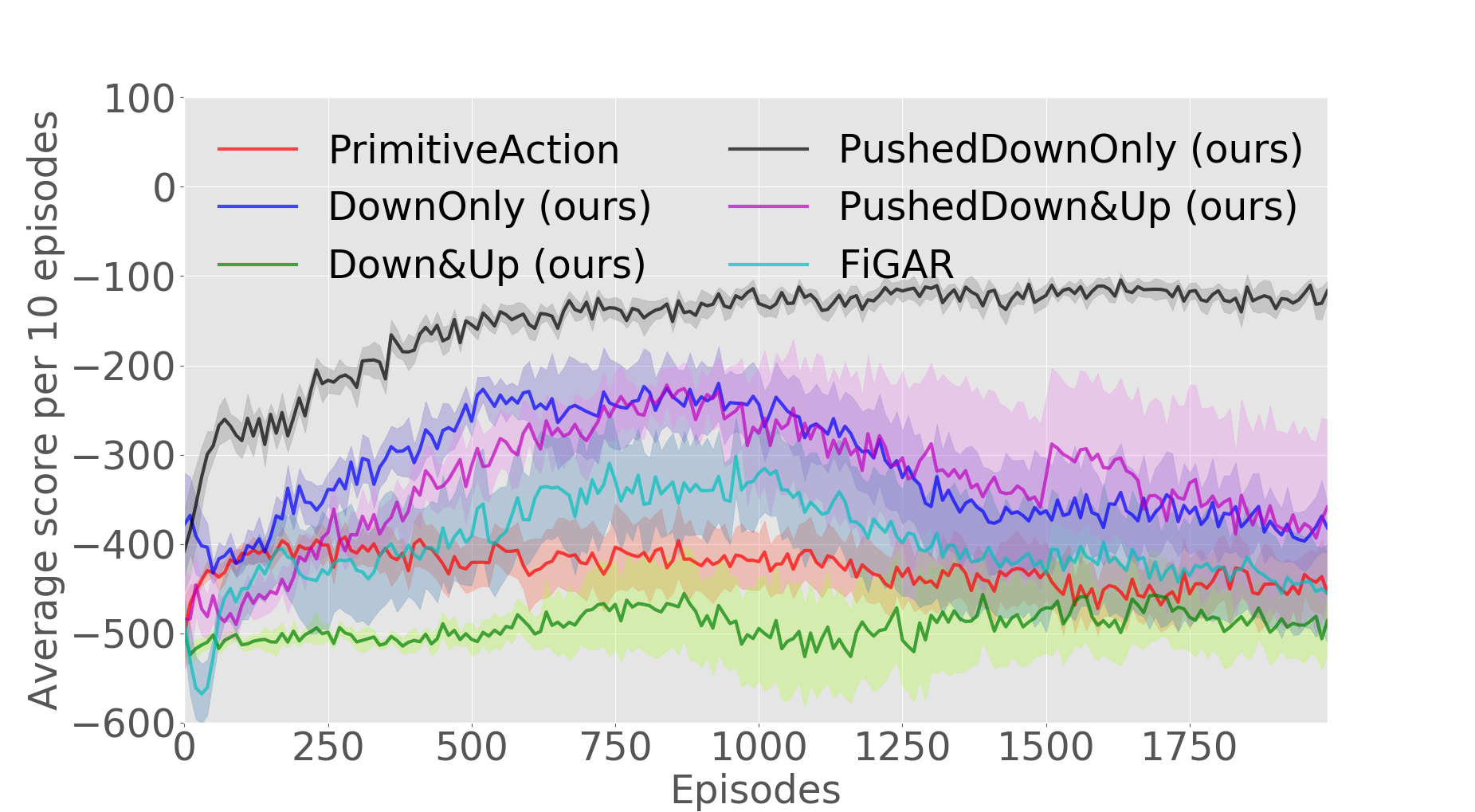}
        \subcaption{Comparison among different actions}
        \label{fig:cont_maze_different_actions}
    \end{minipage}    
    \begin{minipage}{0.5\hsize}
        \includegraphics[width=\linewidth]{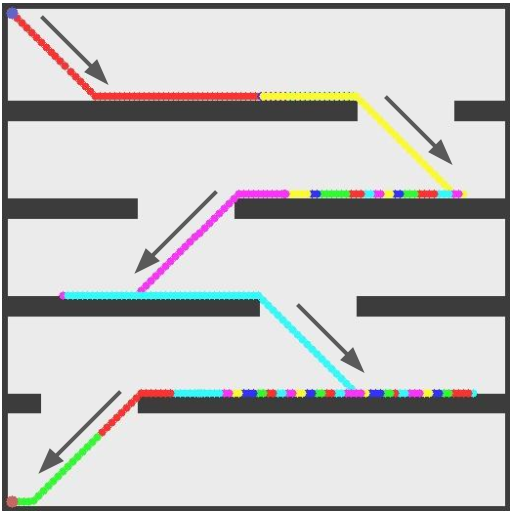}
        \subcaption{Example trajectories of macro actions. Color change indicates change in macro action}
        \label{fig:sample_maze}
    \end{minipage}
    \caption{Results of Maze}
    \label{fig:cont_maze_result}
\end{figure}

With the expert demonstrations generated in Base, we used FaMARL in Maze. We used $\beta=50$. Among FaMARL with macro actions acquired from expert demonstrations of PushedDownOnly, PPO with primitive actions, and FiGAR,  FaMARL performed best for this task and other two algorithms failed to converge (Figure~\ref{fig:cont_maze_different_actions}). It is also obvious that the choice of macro action is critical. While PushedDownOnly outperformed the primitive action, other macro actions could not complete the task. Because PushedDownOnly does not contain any demonstrated actions of moving upwards, this can dramatically diminish the action space to search. On the other hand, Down\&Up is similar to just repeatedly moving one direction, which was not sufficient for completing the task.

\begin{figure}
    \centering
    \begin{minipage}{0.495\hsize}
        \includegraphics[width=\linewidth]{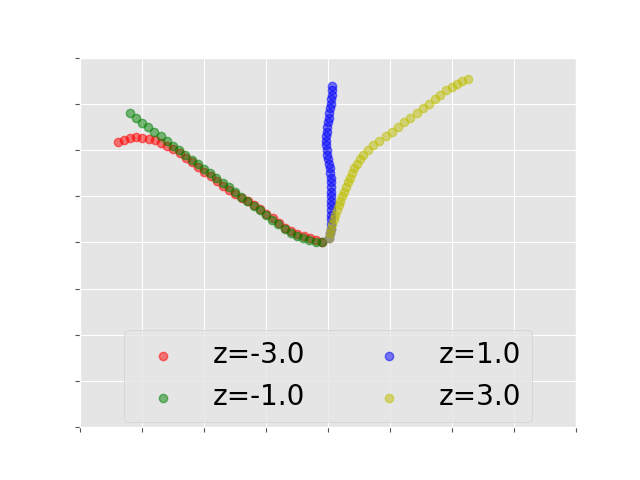}
        \subcaption{(3,1): Node that learned factor}
        \label{fig:2_0}
    \end{minipage}
    \begin{minipage}{0.495\hsize}
        \includegraphics[width=\linewidth]{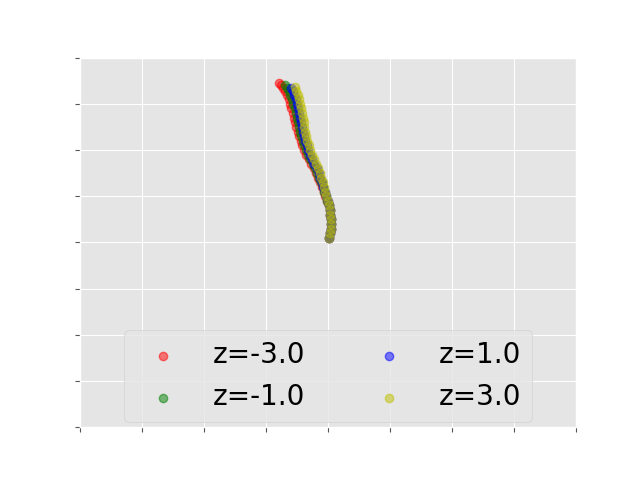}
        \subcaption{(2,1): Node that did not learn any factor}
        \label{fig:1_0}
    \end{minipage}
    \caption{Examples of latent traversal on (Ladder, Index of z) of Down\&Up} \label{fig:latent_traversal}
\end{figure}

Figure~\ref{fig:latent_traversal} shows visualized example trajectories of latent traversal for Down\&Up. Latent traversal is a technique that shifts only one latent variable and fixes the other variables for observing the decoded output from the modified latent variables. If disentangled factor representation is acquired, the output shows meaningful changes. Otherwise, changes are not distinguishable. Also, if the number of latent variables exceeds that of factors that form the sequence of actions, only some of the latent variables acquire factors and the others show no changes when traversed.
Figure~\ref{fig:2_0} shows that the $1$st variable of the $3$rd ladder changed to $\langle -3.0, -1.0, +1.0, +3.0 \rangle$. This changed the direction of the agent's trajectory, while Figure~\ref{fig:1_0} shows no change. This result indicates that FAVAE learns the disentangled representation of a given sequence of actions.

\begin{figure}
    \centering
    \begin{minipage}{0.9\hsize}
        \includegraphics[width=\linewidth]{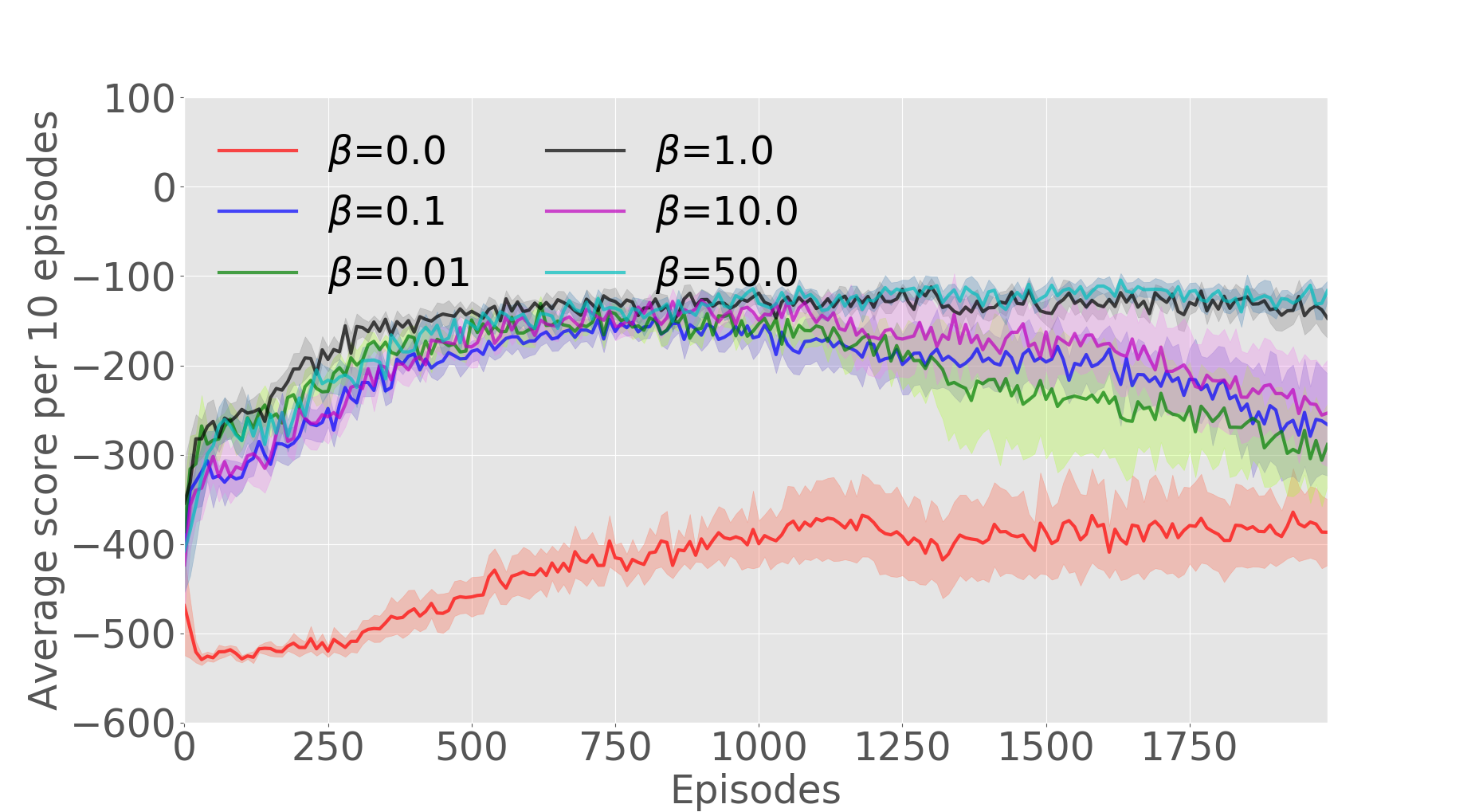}
        \subcaption{Comparison among different $\beta$ on PushedDownOnly}
        \label{fig:cont_maze_different_beta}
    \end{minipage}
    
    \begin{minipage}{0.9\hsize}
        \includegraphics[width=\linewidth]{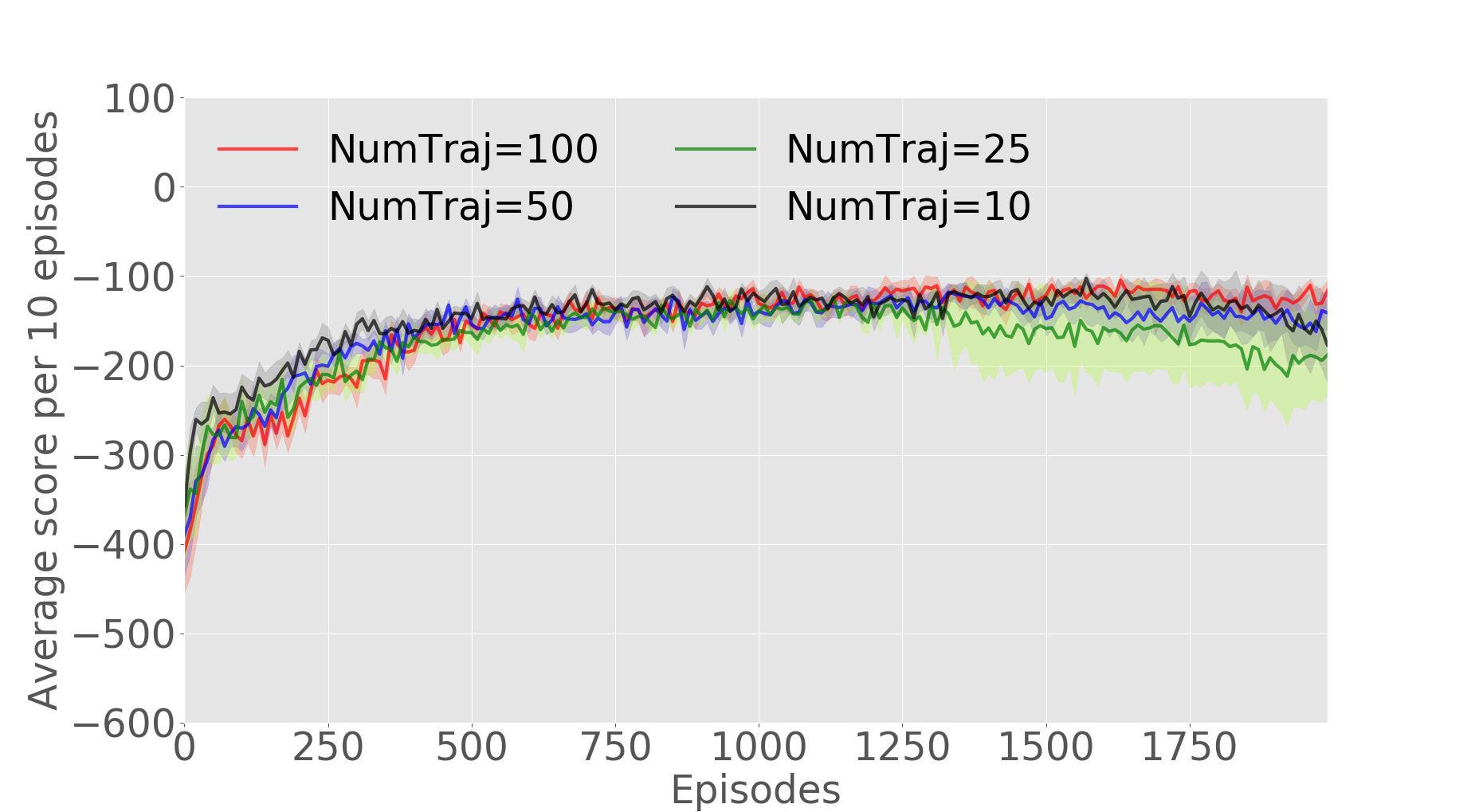}
        \subcaption{Comparison among different numbers of expert trajectories of PushedDownOnly}
        \label{fig:cont_maze_different_expert}
    \end{minipage}
    \caption{Comparison between different $\beta$ and numbers of expert trajectories in Maze}
    \label{fig:cont_maze_comparison}
\end{figure}

Comparison among different $\beta$ of equation~\ref{eq:beta_vae} and numbers of expert trajectories are shown in Figure~\ref{fig:cont_maze_comparison} using PushedDownOnly. Figure~\ref{fig:cont_maze_different_beta} illustrates the experiment with different $\beta$. FAVAE did not learn factors in a disentangled manner when $\beta$ was low. The entangled latent variables of macro actions severely deters matching the state space with macro action space for an optimal policy because the latent space, which actually matches with the state space, is distorted. On ContinuousWorld, we found that $\beta \geq 1.0$ is enough to complete Maze.
Figure~\ref{fig:cont_maze_different_expert} illustrates the experiment with different numbers of expert trajectories. Even though we used 100 expert trajectories across all experiments, the number of trajectories did not impact the performance of FaMARL.

\subsection{RobotHand}


RobotHand has four degrees of freedom (DOFs), i.e., moving along the x axis, moving along the y axis, rotation, and grasping operation. The entire environment was built with Box2D \href{https://box2d.org/}{https://box2d.org/} and rendered with OpenGL \href{https://www.opengl.org/}{https://www.opengl.org/}. 
Similar to Base task at ContinuousWorld, Base at RobotHand, which is a pegging task, provides 100 expert demonstrations to learn disentangled macro actions. And the target tasks Reaching and BallPlacing are completed with the acquired macro actions. We used $\beta=0.1$ on this environment.

\begin{figure}
    \centering
    \begin{minipage}{0.3\hsize}
        \includegraphics[width=\linewidth]{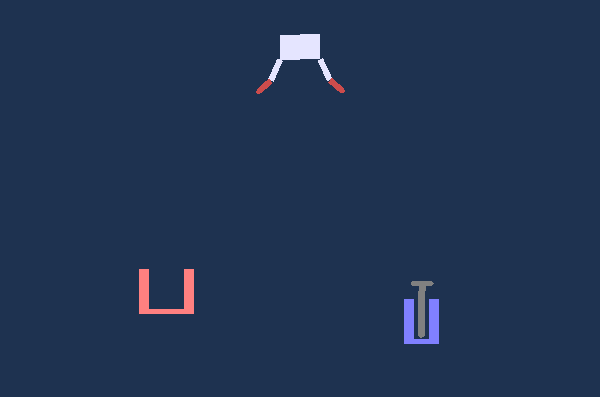}
        \subcaption{Base}
        \label{fig:robot_base}
    \end{minipage}
    \begin{minipage}{0.3\hsize}
        \includegraphics[width=\linewidth]{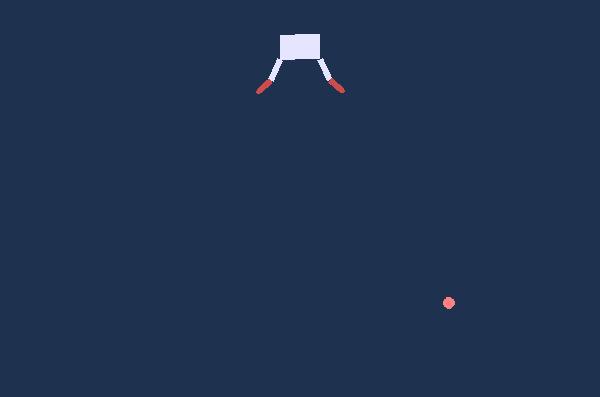}
        \subcaption{Reaching}
        \label{fig:robot_reaching}
    \end{minipage}
    \begin{minipage}{0.3\hsize}
        \includegraphics[width=\linewidth]{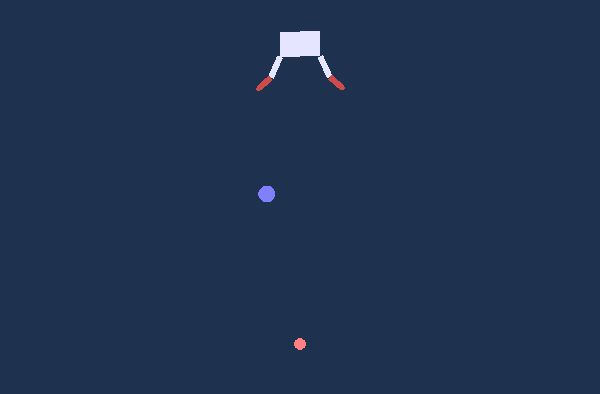}
        \subcaption{BallPlacing}
        \label{fig:robot_ball_placing}
    \end{minipage}
    \caption{RobotHand tasks} \label{fig:robot_tasks}
\end{figure}

Base (Figure~\ref{fig:robot_base}) is a pegging task. In Base, the robot moves a rod from a blue basket to a red one. We chose this task because the pegging task is complex enough to contain all macro actions that might be used in target tasks.

Reaching (Figure~\ref{fig:robot_reaching}) is a simple task. The robot hand has to reach for a randomly generated goal position (red) as fast as possible. To make this task sufficiently difficult, we used a sparse reward setting in which the robot hand only receives a positive reward of +100 for reaching the goal position within a distance of 0.5 m; otherwise there is a time penalty of -1.

In BallPlacing (Figure~\ref{fig:robot_ball_placing}), the robot hand has to carry the ball (blue) to the goal position (red). The ball is initialized at random positions within a certain range, and the goal position is fixed. The reward is defined by $-|b-g|$ where $b$ is the position of the ball and $g$ is the position of the goal. An episode ends when the ball hits the edges or reaches the goal position within a distance of 0.5 m. An additional reward of +200 is given when the ball reaches the goal.


\begin{figure}
    \centering
        \begin{minipage}{0.9\hsize}
        \includegraphics[width=\linewidth]{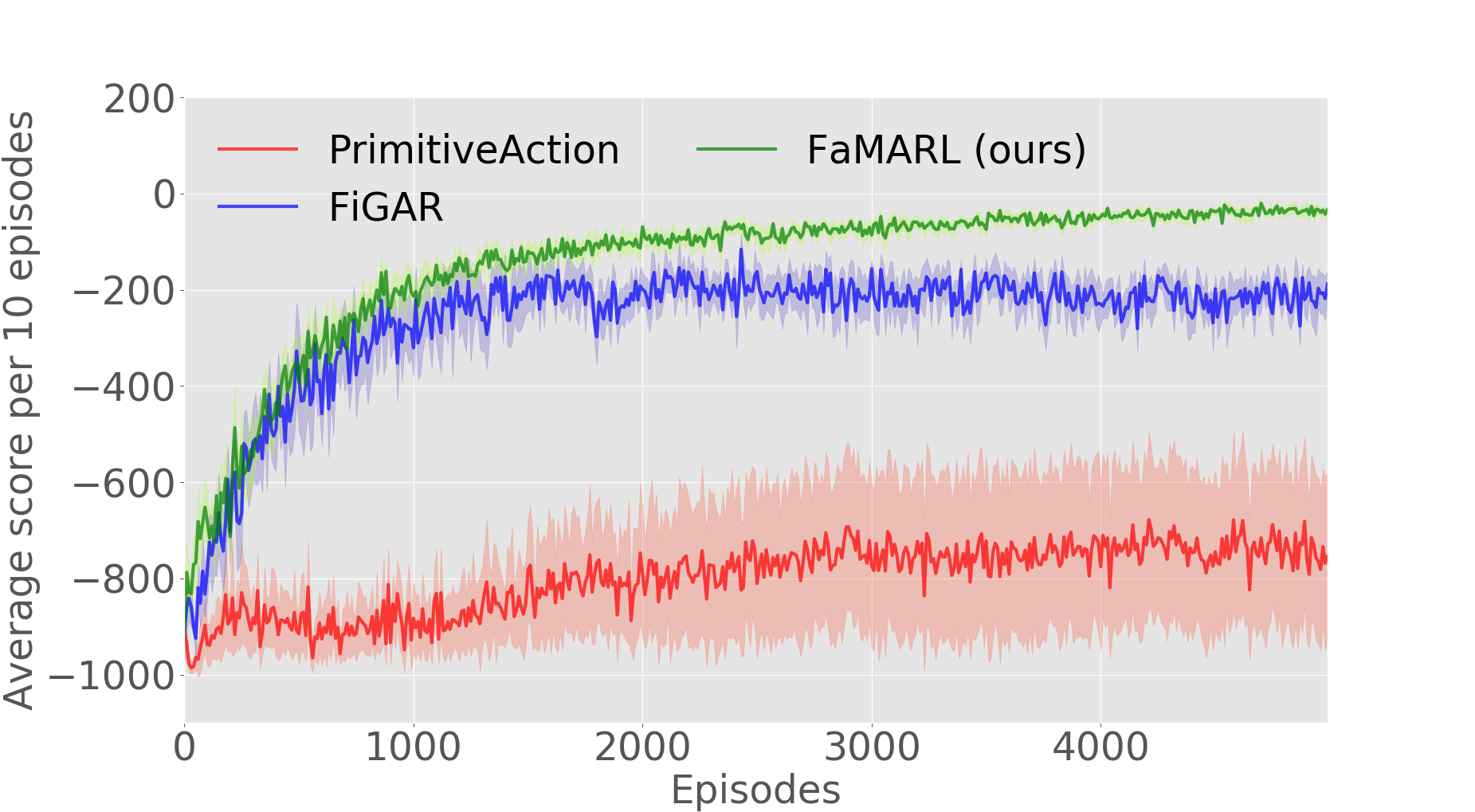}
        \subcaption{Reaching}
        \label{fig:robot_hand_reaching}
    \end{minipage}
    \begin{minipage}{0.9\hsize}
        \includegraphics[width=\linewidth]{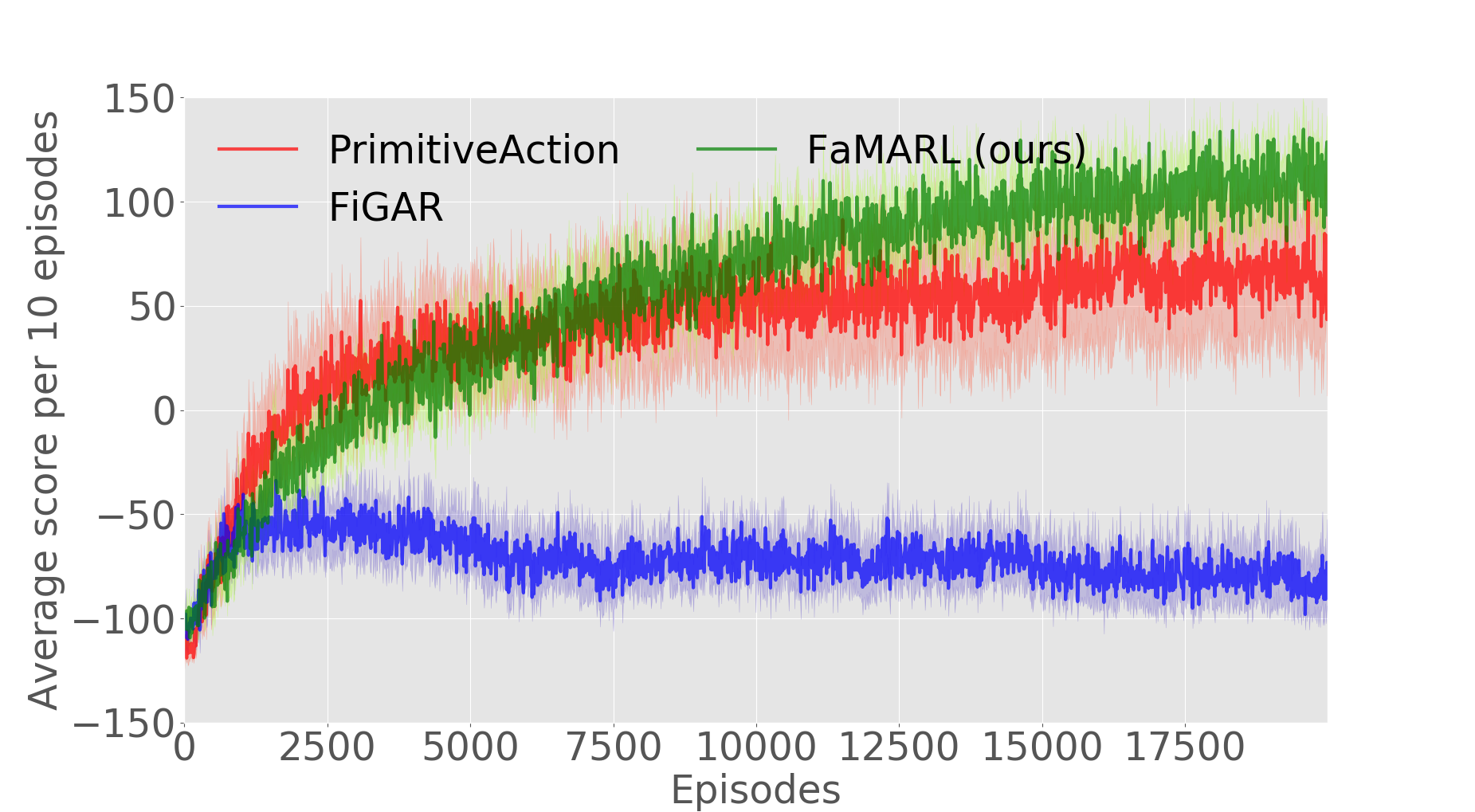}
        \subcaption{BallPlacing}
        \label{fig:robot_hand_ball_placing}
    \end{minipage}
    \caption{Comparison of FaMARL, PPO with primitive actions, and FiGAR in RobotHand tasks}
    \label{fig:robot_hand_result}
\end{figure}

Figure~\ref{fig:robot_hand_result} is a comparison of FaMARL, PPO with primitive actions, and FiGAR on both Reaching and BallPlacing. PPO with primitive actions and FiGAR respectively failed to learn Reaching and BallPlacing, while FaMARL successfully learned both tasks. Because the reward of Reaching is sparse, using primitive actions fails to find rewards. on the other hand, even though the reward of BallPlacing is not sparse, it requires precisely controlling a ball to the goal., FiGAR, which repeats the same primitive actions a number of times, could not precisely control the ball. FaMARL is the only algorithm that completed both tasks.


\begin{figure}
    \centering
    \begin{minipage}{0.9\hsize}
        \includegraphics[width=\linewidth]{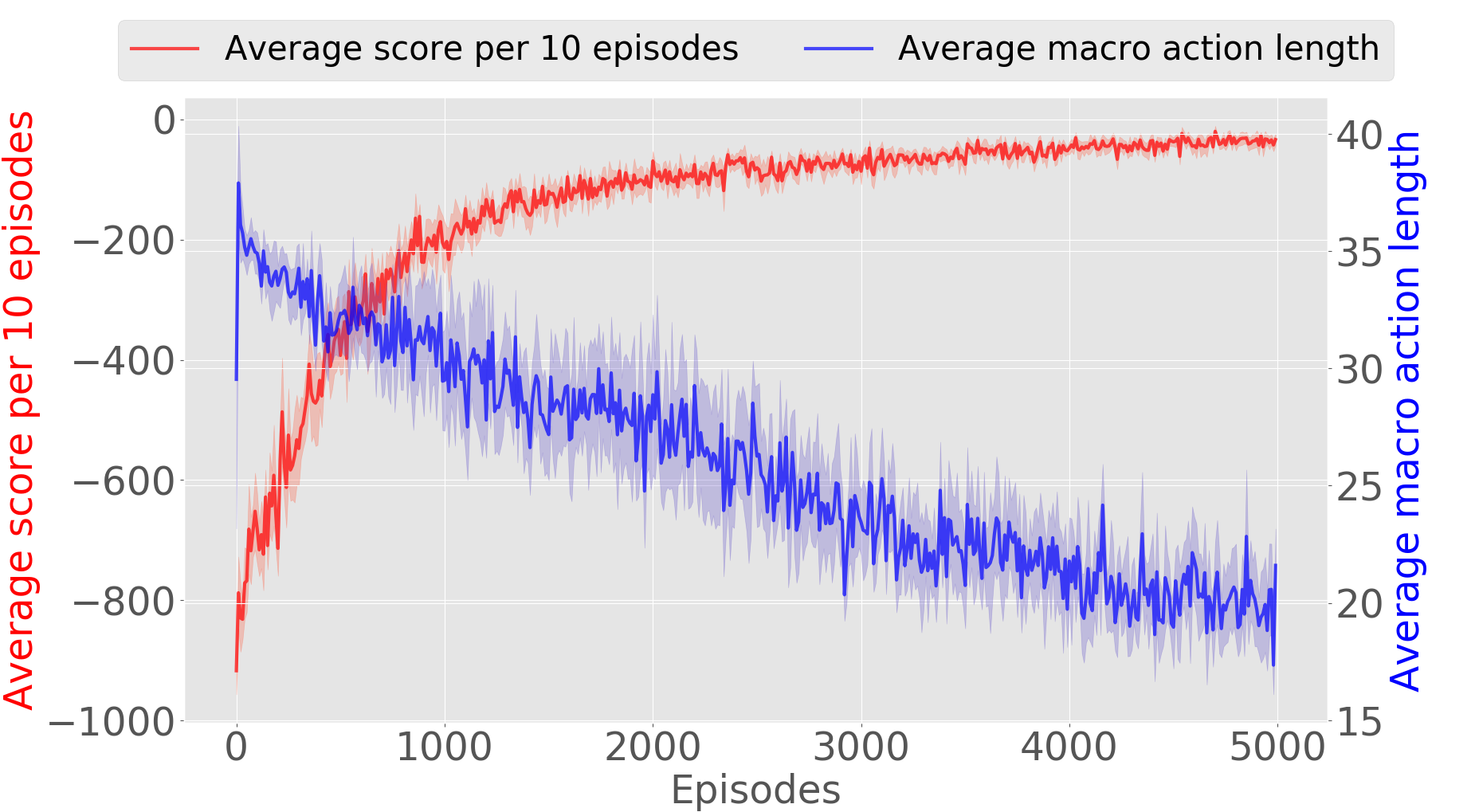}
        \subcaption{Reaching with time penalty}
        \label{fig:action_len_not_time_opt}
    \end{minipage}
    \begin{minipage}{0.9\hsize}
        \includegraphics[width=\linewidth]{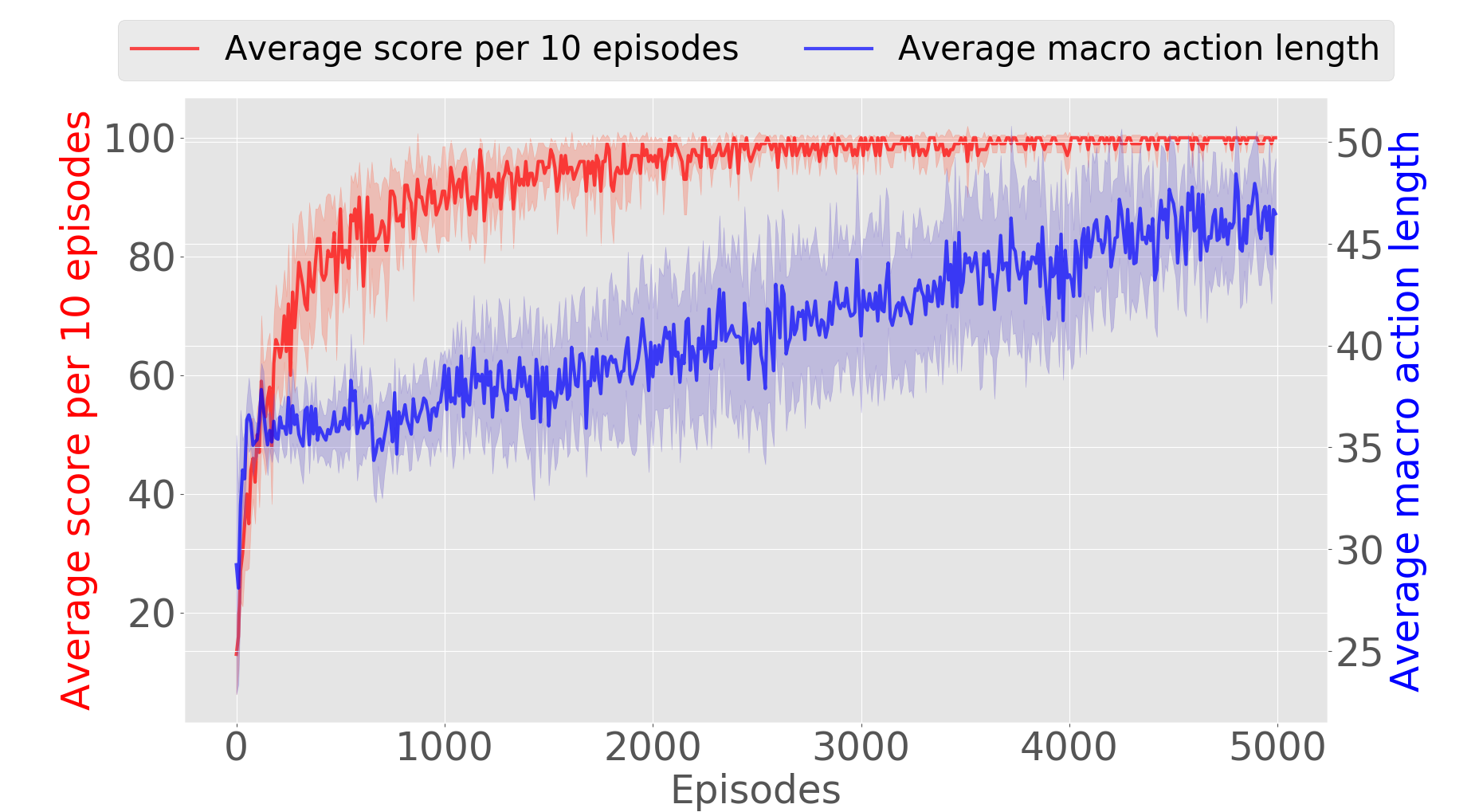}
        \subcaption{Reaching without time penalty}
        \label{fig:action_len_time_opt}
    \end{minipage}
    \caption{Average macro action length and rewards in Reaching with/without time penalty} \label{fig:robot_macro_action_len}
\end{figure}

It should be noted that in the RobotHand experiments FaMARL optimized its behavior by shortening macro actions, while fully using the advantages of exploring with macro actions. In Reaching, the average length of macro actions gradually diminished (Figure~\ref{fig:action_len_time_opt}). However, when time penalty (in Reaching, time penalty of -1 was added to the reward at every time step) is eliminated, the length of a macro action did not diminish (Figure~\ref{fig:robot_macro_action_len}). This is because the agent did not need to optimize its policy in accordance with speed. A macro action can be inefficient in optimizing policy compare to a primitive action because the optimal policy for the task may not match macro actions, but a suboptimal policy will. That is why FaMARL gradually uses primitive-like actions (macro actions with lengths of 1~3) instead of keeping macro actions with dozens of primitive actions.



\section{Limitations of FaMARL}

FaMARL exhibits generally better scores than using primitive actions. However, there are limitations with FaMARL.



\subsection{Lack of feed-back control}
Searching on macro actions instead of primitive actions facilitates searching on the action space in exchange for fast response to unexpected changes in state. We failed to train BipedalWalker-v2\footnote{\href{https://gym.openai.com/envs/BipedalWalker-v2/}{https://gym.openai.com/envs/BipedalWalker-v2/}} with FaMARL based on the expert demonstration at BipedalWalker-v2. Because a bipedal-locomotion task requires highly precise control for balancing induced from instability of the environment; thus, diminishing the search space by macro actions in exchange for faster response was not adequate.

\subsection{Compatibility of macro actions with task}
Figure~\ref{fig:cont_maze_result} shows that the type of macro actions is critical. If the targeted task does not require the macro actions that are abstracted from expert demonstrations, FaMARL will easily fail because the actions an optimal policy requires are not present in the acquired macro actions. Thus, choosing appropriate expert demonstrations for a targeted task is essential for transferring macro actions to target tasks.

\section{Discussion}
We proposed FaMARL, an algorithm of using expert demonstrations to learn disentangled latent variables of macro actions to search on these latent spaces instead of primitive actions directly for efficient search. FaMARL exhibited higher scores than other reinforcement learning algorithms in tasks that require extensive iterations of search when proper expert demonstrations are provided. This is because FaMARL diminishes the searching space based on acquired macro actions. We consider this a promising first step for practical application of macro actions in reinforcement learning in a continuous actions space. However, FaMARL could not complete a task that requries actions outside of macro actions.
the tasks that need actions outside of restricted searching space cannot be solved. Possible solutions include searching optimal policy with both macro actions and primitive actions.


\clearpage

\bibliographystyle{named}
\bibliography{ijcai19}

\end{document}